\documentclass[final,12pt]{clear2025} % Include author names

\title[DVAE-based Hindsight to Learn the Causal Dynamics]{Dynamical-VAE-based Hindsight to Learn the Causal Dynamics of Factored-POMDPs}
\usepackage{times}
\usepackage{hyperref} % For clickable references
\usepackage{array}
\usepackage{booktabs} % For better table formatting
\usepackage{multirow} % For merging rows
\usepackage{caption}  % For table caption formatting
\usepackage{colortbl} % For coloring table cells
\usepackage{xcolor,enumitem}   % For defining custom colors
\usepackage{makecell}
\definecolor{lightgray}{gray}{0.9}
\definecolor{beige}{rgb}{0.96, 0.96, 0.86}
\definecolor{lighttan}{rgb}{0.94, 0.87, 0.8}
\definecolor{lavender}{rgb}{0.9, 0.9, 0.98}
\definecolor{ivory}{rgb}{1.0, 1.0, 0.94}

\SetKwInput{KwParams}{Parameters}

\newcounter{assump}

\allowdisplaybreaks

\clearauthor{
 \Name{Chao Han} \Email{c.han@sheffield.ac.uk}\\
 \addr School of Computer Science, The University of Sheffield, UK \\
 \Name{Debabrota Basu} \Email{debabrota.basu@inria.fr}\\
 \addr Equipe Scool, Univ. Lille, Inria, CNRS, Centrale Lille, UMR 9189 CRIStal, France \\
 \Name{Michael Mangan} \Email{m.mangan@sheffield.ac.uk}\\
 \addr School of Computer Science, The University of Sheffield, UK \\
 \Name{Eleni Vasilaki} \Email{e.vasilaki@sheffield.ac.uk}\\
 \addr School of Computer Science, The University of Sheffield, UK \\
 \Name{Aditya Gilra} \Email{aditya.gilra@cwi.nl}\\
 \addr Machine Learning group, Centrum Wiskunde \& Informatica, Amsterdam, Netherlands \\
 \addr School of Computer Science, The University of Sheffield, UK
 }

\begin{document}

\maketitle

\begin{abstract}%
Learning representations of underlying environmental dynamics from partial observations is a critical challenge in machine learning. In the context of Partially Observable Markov Decision Processes (POMDPs), state representations are often inferred from the history of past observations and actions. We demonstrate that incorporating future information is essential to accurately capture causal dynamics and enhance state representations. To address this, we introduce a Dynamical Variational Auto-Encoder (DVAE) designed to learn causal Markovian dynamics from offline trajectories in a POMDP. Our method employs an extended hindsight framework that integrates past, current, and multi-step future information within a factored-POMDP setting. Empirical results reveal that this approach uncovers the causal graph governing hidden state transitions more effectively than history-based and typical hindsight-based models.
%Learning a representation of the underlying dynamics, i.e. model, of an environment from partial observations is widely useful, from model-based Reinforcement Learning (RL) to scientific modelling. In the online RL setting, in a Partially Observable Markov Decision Process (POMDP), generative state representations are typically learned from a history of observations and actions. Here, we derive that future information is also required in learning the causal dynamics and state representations of the environment.
%While the future steps are unavailable for online RL, these could be available for offline model learning, with the model then used for model-based online RL. 
%We design a Dynamical Variational Auto-Encoder (DVAE) based algorithm to learn underlying causal Markovian dynamics from offline trajectories of a POMDP. We deploy our algorithm in a factored-POMDP, to demonstrate cases where our extended hindsight (involving 1-step past, current and multi-step future), can identify hidden variables better compared to history-based (involving past and current) and typical hindsight-based (involving only current and 1- or multi-step future) approaches. Empirical evaluations also demonstrate that our algorithm more effectively identifies the underlying causal graph of the transitions of the hidden state representations.
\end{abstract}


\section{Introduction}
%Learning a model of the environment enables sample-efficient reinforcement learning (RL) \citep{wang2019benchmarking, moerland2023model}. We would like to learn the true underlying Markovian model of the environment. Current self-predictive state representation approaches ensure that we learn a Markovian transition model \citep{ni_bridging_2024}. However, in a partially observable setting, the full underlying state is not available. Rather, most approaches construct a belief state based on the history of previous state-action observations as a proxy for the true underlying state. Thus, for online learning, the POMDP is converted into an MDP by constructing an approximate belief state from the history of past states and actions \citep{astrom_optimal_1965, subramanian_approximate_2022}. 

Accurately learning the underlying dynamics of an environment is essential for developing models that can reliably predict future states, particularly in partially observable settings \citep{wang2019benchmarking, moerland2023model}. Existing self-predictive approaches to state representation aim to learn a Markovian transition model \citep{ni_bridging_2024}. However, in partially observable contexts, the true underlying state remains hidden, making it necessary to construct an approximate belief state from prior state-action histories as a proxy for the latent state. This approach effectively reformulates the Partially Observable Markov Decision Process (POMDP) as a Markov Decision Process (MDP) that depends solely on past observations and actions to approximate the full state information \citep{astrom_optimal_1965, subramanian_approximate_2022}. Such an approach may, in general, only lead to an approximation of the true MDP.

%This MDP formed by a history of states and actions may only be an approximation of the underlying MDP. While in an online setting, the agent has access to only the past information; in offline RL or model learning, both the past and the future of any time step are typically available. The question arises if we can identify the generating MDP better by considering both the past and the future. By maximizing the log-likelihood of the full trajectories of observations and actions, we derive, using the formalism of Dynamical Variational Auto-Encoders \citep{girin2020dynamical}, which aspects of the past and the future are necessary to decode unobservable variables at each time step. We utilize the Reparametrization Lemma \cite{} to distinguish unobserved variables that appear deterministically versus those that can be represented as exogenous noise, independent at each time step, that render the transitions stochastic. At each step, we require the 1-step past, current, and all future information to reconstruct the deterministic unobserved variables, hereafter called hidden variables. Earlier work  used only the current and 1-step future as hindsight \citep{jarrett_curiosity_2023}. By contrast, we call our latent identification using 1-step past, current and future as `non-myopic hindsight'.

In online settings, the agent is limited to past information alone, but in offline RL or model learning, both past and future data around each time step are accessible. This availability raises the question of whether combining both past and future information can improve our ability to identify the generating MDP. By maximizing the log-likelihood of complete trajectories of observations and actions, we leverage the formalism of Dynamical Variational Auto-Encoders (DVAE) \citep{girin2020dynamical} to determine which elements of the past and future are essential for decoding unobservable variables at each time step. We separate unobservable variables into deterministic hidden ones, and using the Reparameterization Lemma \citep{buesing_woulda_2018}, into exogenous stochastic ones. We find that the 1-step past (including bootstrapped hidden), present, and future observables and actions are needed to accurately reconstruct deterministic unobserved hidden variables. We term our approach ``DVAE-based hindsight'' to contrast it with prior hindsight methods for latent identification that utilized only the present and 1-step future \citep{jarrett_curiosity_2023}.

We utilize Causal Dynamical Learning (CDL) \citep{wang2022causal}, employing Conditional Mutual Information (CMI), to learn a causal transition graph of the environment. The stationary Markovian transition model can be represented as a Directed Acyclic Graph (DAG), mapping the Markovian states and action at time step $t$ to the Markovian states at $t+1$. We extend CDL to a partially observable setting by learning to identify deterministic hidden variables and constructing the causal transition graph, combining the DVAE and CDL approaches in an end-to-end framework. We demonstrate the effectiveness of our approach against history-based \citep{littman2001predictive, baisero2020learning, ni_bridging_2024} and earlier hindsight-based methods \citep{jarrett_curiosity_2023}, in a factored-POMDP setting \citep{oliehoek2021sufficient} which highlights the advantages of our method.

%We utilize Causal Dynamical Learning (CDL) \citep{wang2022causal}, employing 'Conditional Mutual Information' (CMI), to learn a causal transition graph of the environment. A stationary Markovian transition model of the environment can be represented as a causal graph, that is a directed acyclic graph (DAG) from the Markovian states and action at any time step $t$ to the Markovian states at time step $t+1$. We extend the CDL approach to our partially observed setting, by identifying the hidden variables and learning the causal transition graph, end-to-end combining the DVAE and CDL approaches. 

%We demonstrate the effectiveness of our approach compared to history-based and earlier hindsight-based approaches in a simple factored-POMDP setting \citep{oliehoek2021sufficient}, since this brings out our arguments explicitly and clearly.

%We prove which aspects of the past vs future observations are needed to learn the underlying causal transition model, depending on specific conditions of causal connections and noise sources. 

\section{Preliminaries and Problem Formulation}
\subsection{Partially Observable Markov Decision Processes (POMDPs)}
A Markov Decision Process (MDP) in the context of reinforcement learning is defined by a tuple $(S, A, T_a, R_a)$, where $S$ is the set of states, $A$ the set of actions, $T_a(s’|s)$ the probability of transitioning from state $s$ to $s’$ under action $a$, and $R_a(s’, s)$ the reward received for this transition. However, many real-world systems or environments are only partially observable. It is typically assumed that there exists an underlying or generating MDP that gives rise to a Partially Observable Markov Decision Process (POMDP) $(S, A, T_a, R_a, \Omega, O)$, where the states are not directly observable. Instead, we observe elements $o$ from a set $\Omega$, governed by conditional probabilities $O(o|s)$. A POMDP can be converted into an MDP by relying solely on the history of observations and actions \citep{astrom_optimal_1965}. This approach forms the basis for using a sequence of past observations (or their representation) and actions as a proxy, or belief state, for the environment’s current state \citep{subramanian_approximate_2022}.

\subsection{Problem formulation: Learning the causal dynamics underlying a factored-POMDP}
\begin{figure}[t!]
\centering
\includegraphics[width=0.95\textwidth]{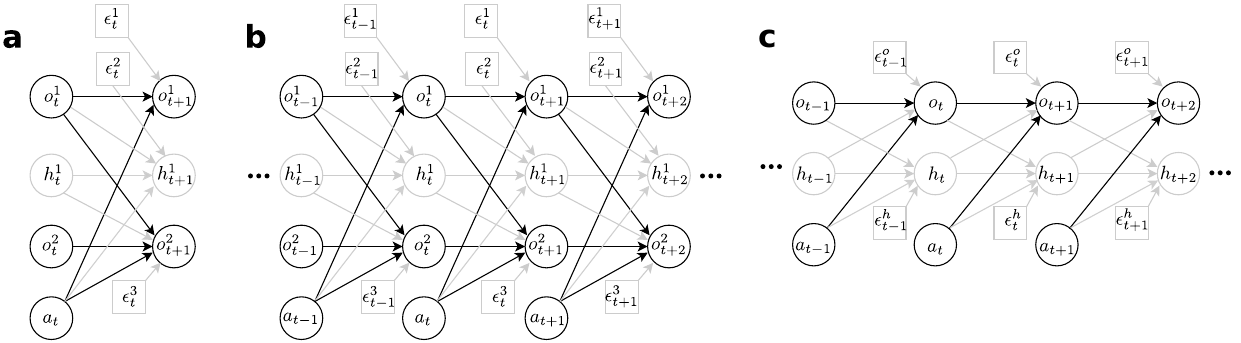}
\caption{\textbf{(a)} The stationary transition model of a factored-POMDP is shown as a Structural Causal Model (SCM) from time step $t$ to $t+1$. The factored states are represented as circle nodes, which are deterministic as per Eq. \eqref{eqn:anm}. They can be either observed (black) or hidden (gray). Gray squares represent unobserved exogenous (i.e. no parents) stochastic nodes. The arrows connecting nodes represent directed causal edges from parents to children. The connectivity of the deterministic nodes is only an example. \textbf{(b)} The stationary transition model can be unrolled over time, by repeating the graph in panel \textbf{(a)}. over multiple time steps, to obtain a SCM for a full trajectory. \textbf{(c)} We collect hidden factored states into vector $h$, and observable factored states into vector $o$ while maintaining the underlying causal model. This is the general SCM for any factored-POMDP.}
\label{fig:graph_model} \vspace*{-1em}
\end{figure}

Our objective is to learn the underlying state transitions and associated causal graph (represented in Figure \ref{fig:graph_model}) from offline data in a factored-POMDP environment.  A factored-POMDP \citep{oliehoek2021sufficient} allows us to focus on learning the underlying transition function and graph, without additional details of representation learning.

\noindent\textbf{Definition 1 (Factored-POMDP)} \citep{oliehoek2021sufficient}.  A factored partially observable Markov decision process is defined as a tuple $\langle S, O, H, A, T, R, \bar{O}\rangle$ where:
\begin{itemize}[leftmargin=*,itemsep=-0.4em,topsep=0pt]
    \setlength\itemsep{-0.4em}
    \item the state space $S$ is spanned as $S=S^1 \times \cdots \times S^{d_S}$ (each state variable $S^k$ is called a factor), such that every state $s \in S$ is a $d_S$-dimension vector $s=( s^1, \ldots, s^{d_S})$.
    \item the space of observed states $O \subseteq S$ is denoted as  $O=O^1 \times \cdots \times O^{d_O}$ with $d_O \leq d_S$.
    \item the space of hidden states $H \subseteq S$ is spanned as  $H=H^1 \times \cdots \times H^{d_H}$ with $d_H \leq d_S$. 
    \item $O \cup H=S$, $O \cap H = \emptyset$, such that $s=\left(o, h\right)$.
    \item $A$ is the set of actions $a$.
    \item $T\left(s_{t+1} \mid s_t, a_t\right)= \prod_{j=1}^{d_H} \prod_{i=1}^{d_O} p(h_{t+1}^j | s_t, a_t) p(o_{t+1}^i | s_t, a_t)$ is the transition probability function.
    \item $R\left(s_t, a_t, s_{t+1}\right)$ is the reward function
    \item $\bar{O}\left(o_t \mid s_t\right)$ is the observation probability function that outputs 1 if $o_t \in O$ is subvector of $s_t \in S$ and 0 otherwise.
\end{itemize}
In this factored-POMDP setting, the state \(s\) is represented as a concatenation of observed and hidden states, denoted by \(s = (o, h)\). The state transition probability distribution \(T\) can be factorised as \(T(s_{t+1} | s_t, a_t) = \prod_{j=1}^{d_S} p(s_{t+1}^j | s_t, a_t)\). Consequently, our goal reduces to learning the factored transitions $p(o_{t+1}^j | \{s_t^i\}_{i=1}^{d_S}, a_t)$ for $j = 1, \ldots, d_O$ and $p(h_{t+1}^j | \{s_t^i\}_{i=1}^{d_S}, a_t)$ for $j = 1, \ldots, d_H$.

% using the Reparametrization Lemma
\noindent\textbf{Representing stochasticity in transitions as independent exogenous noise.} Via the Reparameterization Lemma (Appendix B of \cite{buesing_woulda_2018}), we can always reparameterize the stochasticity to be exogenous, and write the probabilistic MDP transition of factored state variables as a Structural Causal Model (SCM)
\begin{align}
    \label{eqn:anm}
    s_{t+1}^i := f_i(\mathbf{PA}_{s^i_{t+1}}, a_t, \epsilon_{t}^i), \quad i = 1, \dots, d_S
\end{align}
where each $f_i$ represents an arbitrary deterministic function. $\mathbf{PA}_{s^i_{t+1}}$ denotes the set of parent state factors at time $t$, of $s^i_{t+1}$, such that there exists an edge from each element $s_{t}^j \in \mathbf{PA}_{s^i_{t+1}}$ to $s_{t+1}^i$ in the transition graph $\mathcal{G}$. Action $a_t$ is represented separately for clarity. %The binary action variable $a_t^i$ represents whether the $i$-th factor is intervened or not at the current step $t$ by affecting its next state. 
The exogenous noise variable $\epsilon_t^i$ for each factor $i$ is jointly independent at each time step $t$, that is $p_{\epsilon_t^1, \dots, \epsilon_t^{d_S}}=\prod_{i=1}^{d_S} p_{\epsilon_t^i}$. This noise variable can be seen as introducing stochasticity in the transitions, such that every $s_{t+1}^i = f_i(\mathbf{PA}_{s^i_{t+1}}, a_t^i, \epsilon_{t}^i)$ is a sample drawn from $p(s_{t+1}^i|\mathbf{PA}_{s^i_{t+1}}, a_t^i)$, for every $\epsilon_{t}^i$, consistent with the reparameterization lemma \citep{buesing_woulda_2018}. Thus, in Fig. \ref{fig:graph_model}, we can represent all stochasticity in transitions with independent exogenous noise nodes.

%Using the reparametrization lemma \citep{buesing_woulda_2018}, we can always write a factored-MDP in the form of \eqref{eqn:anm} with a bijective mapping $f_i$. (See Appendx .... for the proof.) -- may not be bijective

Furthermore, any stochastic factored-MDP can be converted to the factored-POMDP setting by hiding a set of factors $h=( h^1, \ldots, h^{d_H})$ from the agent. From the perspective of an agent, the uncertainty in predicting the next observables $o_{t+1}^i$ from the current observables and action, in such a setting, arises from two sources: the effect of current values of hidden factors $h_t$ and the unobservable stochasticity in the transition encapsulated by the current noise $\epsilon_t^i$. Therefore, if we somehow had access to the current hidden states $h_t$ and the noise $\epsilon_t^i$, then each next state $s_{t+1}^i$ would be deterministically predictable given the current observed states $o_t$ and action $a_t^i$. For our factored-POMDP, similar to examples in real life, both $h_t$ and $\epsilon_{t}^i$ are not observable.

\section{Deriving the algorithm for learning the transition dynamics of factored-POMDPs}
%In this section, we first outline the assumptions made about the underlying transition graph. %that enables our proposed model to identify hidden variables and infer their causal edges over time. %We discuss scenarios where different combinations of past, present, one-step future, or multi-step future data are required for a minimalistic model, using a simple yet non-trivial graph setup as an illustrative example.
%We illustrate the conditions under which we can infer the hidden states at each time step, in particular when do we need the previous, current and future observables and actions to infer the current hidden state, under stochasticity in hidden versus observed state factors.
In subsection \ref{subsec:dvae}, we derive the DVAE-based framework for identifying the transition model using our extended hindsight encoder for hidden factors. In subsection \ref{app:subsec:cmi}, we outline how we estimate the transition graph. In subsection \ref{subsec:modulo_env}, we outline our Modulo environment, an example factored-POMDP to illustrate our results.

\subsection{DVAE for Factored-POMDP}
\label{subsec:dvae}
We aim to maximize the conditional marginal log-likelihood of the observations $o_{1:T}$ given the actions $a_{1:T}$, parameterized by $\theta$, under the true data distribution $p(o_{1:T} | a_{1:T})$:
\begin{equation}
    \label{eq:mllh}
    \max_\theta \mathbb{E}_{p(o_{1:T} | a_{1:T})}\left[\log p_\theta(o_{1:T} | a_{1:T})\right]
\end{equation}
By introducing a variational distribution $q_{\phi}(h_{1:T} | o_{1:T}, a_{1:T})$, parameterized by $\phi$, we can decompose the objective in Eq. \eqref{eq:mllh} as follows (see Appendix~\ref{app:subsec:llh_decom} for derivation):
\begin{align}
    \label{eq:mllh_decom}
    \max_{\theta, \phi} \mathbb{E}_{p(o_{1:T} | a_{1:T})}[\ell_{\mathrm{VLB}}(\theta, \phi ; o_{1: T}, a_{1: T}) + D_{\mathrm {KL}}(q_\phi(h_{1:T} | o_{1:T}, a_{1:T}) \parallel p_{\theta}(h_{1:T} | o_{1:T}, a_{1:T}))] 
\end{align}
Here, $\ell_{\mathrm{VLB}}(\theta, \phi; o_{1: T}, a_{1: T})$ is the variational lower bound (VLB) on the marginal log-likelihood, serving as a lower bound due to the non-negativity of the KL divergence term. VLB is defined as:
% \begin{align}
%     \label{eq:vlb_general}
%     \ell_{\mathrm{VLB}}\left(\theta, \phi; o_{1: T}, a_{1: T}\right) 
%     =&\mathbb{E}_{q_\phi\left(h_{1: T} | o_{1: T}, a_{1: T}\right)} \left[ \log p_\theta\left(o_{1: T}, h_{1: T} | a_{1: T}\right) \right. \notag \\
%     &\phantom{\mathbb{E}_{q_\phi\left(h_{1: T} | o_{1: T}, a_{1: T}\right)}\left[\right.} \left.- \log q_\phi\left(h_{1: T} | o_{1: T}, a_{1: T}\right) \right] 
% \end{align}
\begin{align}
    \label{eq:vlb_general}
    \ell_{\mathrm{VLB}}\left(\theta, \phi; o_{1: T}, a_{1: T}\right) 
    =&\mathbb{E}_{q_\phi\left(h_{1: T} | o_{1: T}, a_{1: T}\right)} \left[ \log p_\theta\left(o_{1: T}, h_{1: T} | a_{1: T}\right) - \log q_\phi\left(h_{1: T} | o_{1: T}, a_{1: T}\right) \right] 
\end{align}
Thus, optimizing Eq. \eqref{eq:mllh} reduces to maximizing the expected VLB. In practice, we approximate the expectation of the data distribution $p(o_{1:T} | a_{1:T})$, using observed data sequences. We employ independent and identically distributed (i.i.d.) sampled trajectories from the collected dataset $\mathcal{D}$ to construct a Monte Carlo estimate of the expected VLB, defined as follows:
\begin{align}
    \label{eq:vlb_expected}
    \mathcal{L}_{\mathrm{VLB}}\left(\theta, \phi; o_{1: T}, a_{1: T}\right) = 
    \mathbb{E}_{(o_{1:T}, a_{1:T}) \sim \mathcal{D}} \left[ \ell_{\mathrm{VLB}}\left(\theta, \phi; o_{1: T}, a_{1: T}\right) \right]
\end{align}

Leveraging the Markov property in the transition dynamics, the generative model $p_\theta$ and inference model $q_\phi$ in VLB of Eq. \eqref{eq:vlb_general} can be further decomposed along time indices and state factors as follows (see Appendix~\ref{app:subsec:vlb_dvae} for derivation):
\begin{align}
    \label{eq:inf_gen_decom}
    p_{\theta}(o_{1:T}, h_{1:T} | a_{1:T}) =& \prod_{t=0}^{T-1}\prod_{j=1}^{d_H} \prod_{i=1}^{d_O} p_{\theta_{h}}(h_{t+1}^j | s_t, a_t) p_{\theta_{o}}(o_{t+1}^i | s_t, a_t), \\
    q_\phi(h_{1: T} | o_{1: T}, a_{1: T}) =& \prod_{t=0}^{T-1} \prod_{j=1}^{d_H} q_{\phi}(h_{t+1}^j | h_{t}, o_{t:T}, a_{t:T})\label{eq:latent_decom}
\end{align}
where $o_{t} = (o_{t}^1, \ldots, o_{t}^{d_O})$, $h_{t} = (h_{t}^1, \ldots, o_{t}^{d_H})$ and $s_{t} = (o_{t}, h_{t})$. Note that the decomposed terms $ p_{\theta_{h}}(h_{t+1}^j | s_t, a_t)$ and $p_{\theta_{o}}(o_{t+1}^i | s_t, a_t)$ can be interpreted as the transition probabilities of the $j$-th hidden state and $i$-th observed state, respectively. Similarly, $q_{\phi}(h_{t+1}^j | h_{t}, o_{t}, a_{t})$ serves as the encoder for the $j$-th hidden state, which we refer to as the \textit{DVAE-based hindsight encoder}. 
\begin{remark}[DVAE-based hindsight encoder for inferring the current hiddens]
Eq. \eqref{eq:latent_decom} shows that the joint conditional of the hidden states can be decomposed into $T$ conditionals, each conditioned on 1-step past, current and all future observations and actions, as well as the 1-step past hidden states. The previous hidden states are recursively chained across the $T$ time steps, effectively incorporating the entire past. Thus, the hidden encoder $q_{\phi}(h_{t+1}^j | h_{t}, o_{t:T}, a_{t:T})$ systematically leverages all available information to infer the distribution of hidden states.
\end{remark} \vspace*{-1em}
\begin{remark}[History-based encoder vs. DVAE-based hindsight encoder]
A history-based encoder, which conditions only on past and current observables and actions, cannot fully infer the current hidden, as it depends on an exogenous noise variable that is independent of past and current observations and actions (Fig. \ref{fig:graph_model}).
In contrast, future observations, which depend on the current hidden state and carry this noise information, are utilized by our DVAE-based hindsight encoder.
\end{remark} \vspace*{-1em}
\begin{remark}[Current and 1-step hindsight encoder vs. DVAE-based hindsight encoder]
\label{rem:myopic_enc}
Specifically, Eq. \eqref{eqn:anm} can be rewritten as $o_{t+1}^i := f_i(o_t,h_t^j,a_t, \epsilon_{t}^i)$, for every $i$ and $j$. 
An encoder conditioned on $o_t$, $a_t$, and $o_{t+1}^i$, as in \cite{jarrett_curiosity_2023}, would infer $h_t^j$ by inverting the transition function of the parent of $h_t^j$, i.e., $o_{t+1}^i$. However, the inferred $h_t^j$ would be contaminated with $\epsilon_t^i$. 
In fact, they do not include any hidden states in their environment, encoding only the exogenous noise using current and 1-step future observations and actions. 
In our DVAE-based hindsight encoder, the additional information from the 1-step past, along with further future observations and actions and the bootstrapped 1-step past hidden state, better disentangles the current hidden state from the exogenous noise.
\end{remark}
Substituting Eqs. \eqref{eq:inf_gen_decom} and \eqref{eq:latent_decom} into Eq. \eqref{eq:vlb_general} yields:
\begin{align}
    \label{eq:vlb_factorized}
    \ell_{\mathrm{VLB}}\left(\theta, \phi, \overline{\phi}; o_{1: T}, a_{1: T}\right)
    =&\sum_{t=0}^{T-1} \mathbb{E}_{q_\phi\left(h_{1: t} | o_{1: T}, a_{1: T}\right)}\Bigg[ \sum_{j=1}^{d_O} \log p_{\theta_{o}}(o_{t+1}^j | s_{t}, a_{t}) \notag \\
    &- \sum_{j=1}^{d_H} D_{\mathrm {KL}} (q_{\overline{\phi}}(h_{t+1}^j | h_{t}, o_{t:T}, a_{t:T}) \parallel p_{\theta_{h}}(h_{t+1}^j | s_{t}, a_{t})) \bigg]
\end{align}
Here, $q_{\overline{\phi}}(h_{t+1}^j | h_{t}, o_{t:T}, a_{t:T})$ serves as the target distribution of the next encoded hidden state in the KL divergence term, comparing it to the distribution of the next predicted hidden state $ p_{\theta_{h}}(h_{t+1}^j | s_{t}, a_{t})$. The notation $\overline{\phi}$ denotes the stop-gradient version of $\phi$, which is detached from the computation graph and replaced by a copy of $\phi$ from the previous training step. Using a stop-gradient target in self-predictive representations is common in practice \citep{zhang2020learning, ghugare2022simplifying}, as this technique helps avoid representational collapse \citep{ni_bridging_2024}.

Building upon this, to infer the factor-wise transition graph, we employ the approach from \cite{wang2022causal}. This involves modifying Eq. \eqref{eq:vlb_factorized} to include 3 types of losses — full, masked, and causal. Thus, in addition to the full transition distribution (without masking any input factor), we compute a masked transition distribution by masking a randomly chosen state factor $s^i_t$ or action $a_t$ from each input $\{ s_t, a_t \}$ to the transition model, and also a causal transition distribution by masking out all input factors except for causal parents of $s^j_{t+1}$ identified using the transition graph learned so far (see section \ref{subsec:cmi}). These are used in the 3 loss types, for both the Negative Log-Likelihood (NLL) of observed states and the KL-divergence (KL-div) of hidden states, to yield 6 loss terms:
% \begin{align}
% \label{eq:vlb_final}
%     \ell_{\mathrm{VLB}}\left(\theta, \phi, \overline{\phi}; o_{1: T}, a_{1: T}\right)
%     =&\sum_{t=0}^{T-1} \mathbb{E}_{q_\phi\left(h_{1: t} | o_{1: T}, a_{1: T}\right)}\Bigg[ \sum_{j=1}^{d_O} \big[ \log p_{\theta_{o}}(o_{t+1}^j | s_{t}, a_{t}) \notag \\
%     &+ \log p_{\theta_{o}}(o_{t+1}^j | s_{t} \backslash s_{t}^i, a_{t}) + \log p_{\theta_{o}}(o_{t+1}^j | \textbf{PA}_{o^j}) \big] \notag \\
%     &- \sum_{j=1}^{d_H} \big[ D_{\mathrm {KL}} (q_{\overline{\phi}}(h_{t+1}^j | h_{t}, o_{t:T}, a_{t:T}) \parallel p_{\theta_{h}}(h_{t+1}^j | s_{t}, a_{t})) \notag \\
%     &+ D_{\mathrm {KL}} (q_{\overline{\phi}}(h_{t+1}^j |h_{t}, o_{t:T}, a_{t:T}) \parallel p_{\theta_{h}}(h_{t+1}^j | s_{t} \backslash s_{t}^i, a_{t})) \notag \\
%     & + D_{\mathrm {KL}} (q_{\overline{\phi}}(h_{t+1}^j |h_{t}, o_{t:T}, a_{t:T}) \parallel p_{\theta_{h}}(h_{t+1}^j | \textbf{PA}_{h^j})) \big] \bigg]
% \end{align}
{\small
\begin{align}
\label{eq:vlb_final}
    & \ell_{\mathrm{VLB}}\left(\theta, \phi, \overline{\phi}; o_{1: T}, a_{1: T}\right)
    = \sum_{t=0}^{T-1} \mathbb{E}_{q_\phi\left(h_{1: t} | o_{1: T}, a_{1: T}\right)}\Bigg[ -\sum_{j=1}^{d_O} \big[ \underbrace{-\log p_{\theta_{o}}(o_{t+1}^j | s_{t}, a_{t})}_{\text{Full NLL Loss}} \underbrace{-\log p_{\theta_{o}}(o_{t+1}^j | s_{t} \backslash s_{t}^i, a_{t})}_{\text{Masked NLL Loss}} \notag \\
    & \underbrace{-\log p_{\theta_{o}}(o_{t+1}^j | \textbf{PA}_{o^j_{t+1}})}_{\text{Causal NLL Loss}} \big] - \sum_{j=1}^{d_H} \big[ \underbrace{D_{\mathrm {KL}} (q_{\overline{\phi}}(h_{t+1}^j | h_{t}, o_{t:T}, a_{t:T}) \parallel p_{\theta_{h}}(h_{t+1}^j | s_{t}, a_{t}))}_{\text{Full KL-Div Loss}} \notag \\
    &+ \underbrace{D_{\mathrm {KL}} (q_{\overline{\phi}}(h_{t+1}^j |h_{t}, o_{t:T}, a_{t:T}) \parallel p_{\theta_{h}}(h_{t+1}^j | s_{t} \backslash s_{t}^i, a_{t}))}_{\text{Masked KL-Div Loss}} + \underbrace{D_{\mathrm {KL}} (q_{\overline{\phi}}(h_{t+1}^j |h_{t}, o_{t:T}, a_{t:T}) \parallel p_{\theta_{h}}(h_{t+1}^j | \textbf{PA}_{h^j_{t+1}}))}_{\text{Causal KL-Div Loss}} \big] \bigg]
\end{align}}
Here, $s_{t} \backslash s_{t}^i = \{s^1_t, \ldots, s^{i-1}_t, s^{i+1}_t, \ldots, s^{d_S}_t\}$ denotes the set of all state factors at time $t$ except for the $i$-th factor $s^i_t$. For each $j$, the index $i$ is uniformly sampled from $\{1, \ldots, d_S\}$. The term $\textbf{PA}_{s^j_{t+1}}$ are inferred from the learned transition graph so far using the conditional mutual information between each pair of factors, as discussed in Section \ref{subsec:cmi}.
\begin{figure}[t!]
\centering
\includegraphics[width=0.95\textwidth]{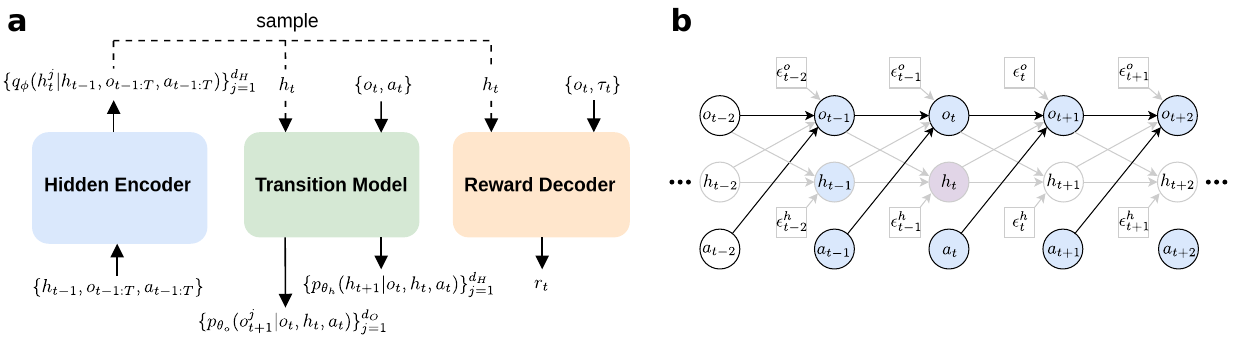}
\caption{\textbf{(a)} Model architecture for computing the objective function in Eq. \eqref{eq:final_loss}. \textbf{(b)} The current hidden state (light purple) is inferred from the hidden states, observables, and action of the previous step, along with the current and all future observables and actions (light blue) within the DVAE-based hindsight encoder.}
\label{fig:encoder_input} \vspace*{-1em}
\end{figure}

Finally, the KL-div term in Eq. \eqref{eq:vlb_final} which matches the distributions between the encoded next hidden state and the predicted next hidden state, can lead to convergence to a trivial constant representation of the hidden state \citep{ni_bridging_2024}. To prevent such degeneration, additional constraints on the hidden representation need to be applied alongside the VLB. Here, we employ a reward predictor parameterized by $\psi$ to condition the encoded hidden representations $h_{1:T}$, trained by minimizing the prediction error:
\begin{align}
    \label{eq:rew_loss}
    \mathcal{L}_{\mathrm{rew}}\left(\phi, \psi; o_{1:T}, \tau_{1:T}, r_{1:T}\right) =
    \mathbb{E}_{\substack{(o_{1:T}, \tau_{1:T}, r_{1:T}) \sim \mathcal{D} \\
    h_{1:T} \sim q_{\phi}(h_{1:T}|o_{1:T}, a_{1:T})}} 
    \left[ \ell_{\mathrm{rew}}\left(\psi; o_{1:T}, h_{1:T}, \tau_{1:T}, r_{1:T}\right) \right]
\end{align}
Here, $\tau_t$ denotes any reward-related variables (e.g., a time-dependent/episodic target) used to predict the reward accurately. $\ell_{\mathrm{rew}}$ can be any supervised loss function; in our experiments, we use cross-entropy loss for categorical rewards.

Combining all components, we obtain the final objective to be minimized. This objective is a weighted sum of the mean VLB from Eqs. \eqref{eq:vlb_expected} and \eqref{eq:vlb_final}, and mean reward loss from Eq. \eqref{eq:rew_loss}, with a weight coefficient $\lambda>0$:
\begin{align}
    \label{eq:final_loss}
    \mathcal{L}_{\mathrm{obj}}\left(\theta, \phi, \overline{\phi}, \psi; o_{1: T}, a_{1: T}, \tau_{1:T}, r_{1:T}\right) = - \mathcal{L}_{\mathrm{VLB}}\left(\theta, \phi, \overline{\phi}; o_{1: T}, a_{1: T}\right) + \lambda \mathcal{L}_{\mathrm{rew}}\left(\phi, \psi; o_{1:T}, \tau_{1:T}, r_{1:T}\right) 
\end{align}

The model architecture depicted in Fig. \ref{fig:encoder_input}a illustrates that every hidden states $h^j_t$ is obtained through temporally recursive sampling from $q_\phi(h_\tau^j | h_{\tau-1}, o_{\tau-1: T}, a_{\tau-1: T})$ for $\tau=1$ to $t$. Then, the hidden sample at each time step $t$ is fed into the transition model and reward decoder to predict next states and reward. The unrolled probabilistic transition graph in Fig. \ref{fig:encoder_input}b highlights the temporal data used as inputs to the DVAE-based hindsight encoder for the hidden states.

\SetKwComment{Comment}{/* }{ */}

\begin{algorithm}[tb]
\caption{Causal Dynamics Learning with Hindsight}
\label{alg:algorithm}
\KwIn{Initial hidden encoder $q_\phi$, initial transition models $p_{\theta_{o}}$ and $p_{\theta_{h}}$, initial reward predictor $R_{\psi}$, and replay buffer $\mathcal{D}$ containing pre-collected data.}
\KwParams{Learning rate $\alpha > 0$, CMI threshold $\delta > 0$, training steps $M$, CMI eval. period $N$.}
\KwOut{Converged hidden encoder $q_{\phi^*}$, transition models $p_{\theta_{o}^*}$, $p_{\theta_{h}^*}$, and graph $\mathcal{G}^*$, reward predictor $R_{\psi^*}$.}

\For{$k = 1$ to $M$ training steps}{
    Update $\mathcal{D}$ and randomly sample a minibatch of $m$ episodes $\left\{o_{1:T}^{(e)}, a_{1:T}^{(e)}, \tau_{1:T}^{(e)}, r_{1:T}^{(e)}\right\}_{e=1}^{m}$\;
    
    Compute the mean objective $\mathcal{L}_{\mathrm{obj}} \left(\theta, \phi, \overline{\phi}, \psi; o_{1:T}^{(1:m)}, a_{1:T}^{(1:m)}, \tau_{1:T}^{(1:m)}, r_{1:T}^{(1:m)}\right)$ using Eq. \eqref{eq:vlb_final}\;

    Update the model parameters;
    \begin{align}
        [\theta_{o}, \theta_{h}, \phi, \psi] &\gets [\theta_{o}, \theta_{h}, \phi, \psi] + \alpha \nabla \mathcal{L}_{\mathrm{obj}} \left(\theta, \phi, \overline{\phi}, \psi; o_{1:T}^{(1:m)}, a_{1:T}^{(1:m)}, \tau_{1:T}^{(1:m)}, r_{1:T}^{(1:m)}\right) \notag \\
        \overline{\phi} &\gets \phi \notag
    \end{align}

    \If{$k \bmod N = 0$}{
        Evaluate $\mathrm{CMI}^{i,j}$ using Eqs. \eqref{eqn:CMI_o} and \eqref{eqn:CMI_h}, and update it with an exponential moving average;
        
        Binarize $\mathrm{CMI}^{i,j}$ to construct $\mathcal{G}$ by checking if $\mathrm{CMI}^{i,j} \geq \delta$;
    }
}
\end{algorithm}

\subsection{Transition Graph Estimation}
\label{subsec:cmi}
The causal dependency of each transition pair $s_t^i \to s_{t+1}^j$ or $a_t \to s_{t+1}^j$ is estimated through conditional mutual information (CMI) \citep{wang2022causal}. During evaluation, the CMI is computed based on two learned transition distributions: the full transition model $p_{\theta}(s_{t+1}^j | s_{t}, a_{t})$, which leverages all state variables and the action to predict the next state of the $j$-th factor, and the masked transition model $p_{\theta}(s_{t+1}^j | s_{t} \backslash s_t^i, a_{t})$, which relies on all state factors except for $s_t^i$ for prediction.

Specifically, when the next state $s_{t+1}^j$ is observable (denoted as $o_{t+1}^j$), the $\mathrm{CMI}^{i,j}$ between $s_t^i$ and $o_{t+1}^j$ given $\{s_{t} \backslash s_t^i, a_{t}\}$ is formulated as:
\begin{equation}
    \label{eqn:CMI_o}
    \begin{aligned}
    I(s_t^i; o_{t+1}^j | s_{t} \backslash s_t^i, a_{t}) =& \mathbb{E}_{s_{t}, a_{t}, o_{t+1}^j \sim \mathcal{D}, q_{\phi}} \left[ \log \frac{p_{\theta_o}(o_{t+1}^j | s_{t}, a_{t})}{p_{\theta_o}(o_{t+1}^j | s_{t} \backslash s_{t}^i, a_{t})} \right] 
    \end{aligned}
\end{equation}
Here, $s_{t}^i$ can be either an observed state or a hidden state sampled from the hidden encoder. The expectation in the CMI is approximated by aggregating transitions from all episodes in a mini-batch.

When the next state $s_{t+1}^j$ is hidden (denoted as $h_{t+1}^j$), the $\mathrm{CMI}^{i,j}$ between $s_t^i$ and $h_{t+1}^j$ conditioned on $\{s_{t} \backslash s_t^i, a_{t}\}$ is given by:
\begin{equation}
    \label{eqn:CMI_h}
    \begin{aligned}
    I(s_t^i; h_{t+1}^j | s_{t} \backslash s_t^i, a_{t}) =& \mathbb{E}_{s_{t}, a_{t} \sim \mathcal{D}, q_{\phi}} \left[ D_{\mathrm {KL}} ( p_{\theta_{h}}(h_{t+1}^j | s_{t}, a_{t}) \parallel p_{\theta_{h}}(h_{t+1}^j | s_{t} \backslash s_{t}^i, a_{t}) ) \right]
    \end{aligned}
\end{equation}
The derivations of Eqs. \eqref{eqn:CMI_o} and \eqref{eqn:CMI_h} are provided in Appendix~\ref{app:subsec:cmi}. Note that for causal dependency between the action and the next state, $a_t \to s_{t+1}^j$, the same CMI formula applies by replacing $s_t^i$ with $a_t$ in the conditioning set, which then becomes $\{s_{t}\}$.

In practice, the existence of an edge in the transition graph, i.e., $s_t^i \to s_{t+1}^j$ or $a_t \to s_{t+1}^j$, is determined by whether the corresponding CMI value $\mathrm{CMI}^{i,j}$ exceeds a predefined threshold $\delta$. The binarized CMI matrix is then applied to select the parents of each next state  in the causal transition losses in Eq. \eqref{eq:vlb_final}, and thus, refines learning of the causal transition dynamics $p_{\theta}(s_{t+1}^j | \textbf{PA}_{s^j_{t+1}})$.

\subsection{Modulo environment: a stochastic, discrete state-action, factored-POMDP}\label{subsec:modulo_env}

\begin{figure}[h!]\vspace*{.5em}
\centering
\includegraphics[width=0.8\textwidth]{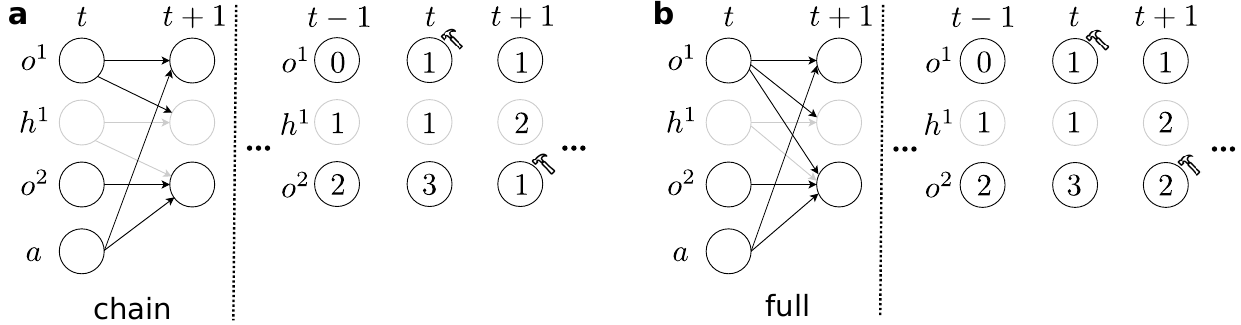}
\caption{Illustration of Modulo environment with different types of transition graphs that have $d_S=3$ and $l=4$. \textbf{(a)} Chain structure. left: ground truth transition graph, right: next states depend on current states and action. \textbf{(b)} Same demonstration for the full structured (lower-triangular adjacency matrix) transition graph. The hammer symbol denotes the action intervened on any of the observed states at each time step.}\label{fig:modulo_env}\vspace*{-.5em}
\end{figure}

Modified from \cite{ke2021systematic}, we construct a probabilistic discrete Factored-POMDP environment,  to examine the performance of our model on inferring the hidden states and underlying transition graph. We called this environment modulo environment as the modulo operator is involved in its transition dynamics defined as $s_{t+1} := ( A s_t + a_t + \epsilon_{t+1} )\ \text{mod}\ l$, 
% \begin{align}
%     \label{eqn:mod_env}
%     s_{t+1} := ( A s_t + a_t + \epsilon_{t+1} )\ \text{mod}\ l
% \end{align}
where $l$ denotes the number of possible discrete values and $A$ is the adjacency matrix of the transition graph $\mathcal{G}$. At time step $t$, each discrete factor $s_t^i$ of the state vector $s_t=( s_t^1, \ldots, s_t^{d_S})^\top$ has values within $\{ 0, \dots, l-1  \}$, the binary element $a_t^i$ of the action vector $a_t=( a_t^1, \ldots, a_t^{d_S})^\top$ represents if the $i$-th factor is intervened or not by setting $a_t^i=1$ or 0 respectively, and the noise vector $\epsilon_t=( \epsilon_t^1, \ldots, \epsilon_t^{d_S})^\top \in E$ is sampled from a jointly independent distribution $p_{\epsilon_t}=\prod_{i=1}^{d_S} p_{\epsilon_t^i}$. Fig. \ref{fig:modulo_env} depicts noise-free transition dynamics with different underlying transition graph structures.

 Our environment satisfies two properties. \textbf{(P1)} For every hidden factor $h_{t}^i$, there exists at least one observable $o_{t+1}^j$, such that $h^i_t \in \mathbf{PA}_{o_{t+1}^j}$. \textbf{(P2)} The transition map $f\equiv\{f_i\}_{i=1}^{d_S}$ in Eq. \eqref{eqn:anm}, for every $a \in A$ and every $\epsilon \in E$, i.e. $(f)_{a,\epsilon}:S\to S$ from any $s_t \in \mathcal{S}$ to $s_{t+1} \in \mathcal{S}$, is bijective, where $s$ is the full state with all observable and hidden factors. 

Indeed, by assuming a version of \textbf{(P2)}, in an environment with only exogenous noise but no hidden factors, we can deterministically infer these exogenous noise variables at $t$, by using a current and 1-step hindsight encoder for the hiddens similar to the latent generator in \cite{jarrett_curiosity_2023}, which learns to invert $f$ using observables and action at current $t$ and observables at 1-step future ${t+1}$. However, \textit{with both hidden factors and exogenous noise, despite these simplifying properties, history-based, and current and 1-step hindsight-based approaches are unable to learn the hidden factor and the graph}, as shown by the following experiments (see also Remark \ref{rem:myopic_enc}).

\section{Experiments demonstrate the effectiveness of DVAE-based hindsight encoder}
\textbf{Environment setting.} We consider a straightforward yet non-trivial setup using the modulo environment: a chain-structured transition graph with $d_S=3$ and $l=4$, with 3 factors: an observable $o^1$, a middle hidden state $h^1$, then an observable $o^2$. The environment includes a stationary discrete noise distribution defined as $p(\epsilon_t^i=-1)=p(\epsilon_t^i=1)=0.05$ and $p(\epsilon_t^i=0)=0.9$ for $i = 1, 2, 3$. The initial hidden state remains fixed across episodes. The principles outlined here can be extended to other graph structures and larger values of $d_S$, as empirically demonstrated later. Specifically, the transition dynamics in this setup are defined as $o_{t+1}^1 := (o_{t}^1 + a^1_t + \epsilon_{t}^1 ) \text{ mod}\ 4$, $h_{t+1}^1 := (o_{t}^1 + h_{t}^1 + \epsilon_{t}^2 )\ \text{mod}\ 4$, and $o_{t+1}^2 := (h_{t}^1 + o_{t}^2 + a^3_t + \epsilon_{t}^3 )\ \text{mod}\ 4$. 
% \begin{equation}
%     \label{eqn:chain3}
%     \begin{aligned}
%         o_{t+1}^1 := (o_{t}^1 + a^1_t + \epsilon_{t}^1 )&\ \text{mod}\ 4\,, \quad
%         h_{t+1}^1 := (o_{t}^1 + h_{t}^1 + a^2_t + \epsilon_{t}^2 )\ \text{mod}\ 4 \\
%         o_{t+1}^2 &:= (h_{t}^1 + o_{t}^2 + a^3_t + \epsilon_{t}^3 )\ \text{mod}\ 4
%     \end{aligned}
% \end{equation}

\noindent\textbf{Baselines and our DVAE encoders.} We compare the performance of 5 different hidden encoders, each learned end-to-end with the same transition model and reward predictor architecture:
% \begin{itemize}[leftmargin=*]
    % \setlength\itemsep{-0.4em}
    (i) history-based encoder, using complete past and current observations and actions (\textbf{History Enc.}): $q_{\phi}(h_{t} | o_{1: t}, a_{1: t})$ parameterized by a forward RNN;
    (ii) Current and 1-step hindsight encoder \citep{jarrett_curiosity_2023}, using current observations and action, and next step future observations (\textbf{Current \& 1-Step Hindsight Enc.}): $q_{\phi}(h_{t} | o_{t: t+1}, a_{t: t+1})$ parameterized by an MLP;
    (iii) Current and full hindsight encoder, using current and all future observations and actions (\textbf{Current \& Full Hindsight Enc.}): $q_{\phi}(h_{t} | o_{t: T}, a_{t: T})$ parameterized by a backward RNN;
    (iv) DVAE-based encoder with 1-step hindsight, using 1-step past (including sampled hidden), current, and 1-step future observations and actions (\textbf{DVAE 1-Step Hindsight Enc.}): $q_{\phi}(h_{t} | h_{t-1}, o_{t-1: t+1}, a_{t-1: t+1})$; and 
    (v) DVAE-based encoder with full hindsight, using 1-step past (including sampled hidden), current, and all future observations and actions (\textbf{DVAE Full Hindsight Enc.}): $q_{\phi}(h_{t} | h_{t-1}, o_{t-1: T}, a_{t-1: T})$.
% \end{itemize}

\noindent\textbf{Implementation details.} We use the Adam optimizer with a step-decayed learning rate $\alpha$. The prediction horizon $T$ is 5, and CMI threshold $\delta$ is 0.03. Details on the neural network parameterization of the hidden encoder, transition model, and reward predictor are provided in Appendix~\ref{app:subsec:nn_paramz}.

\begin{figure}[ht]
\centering
\includegraphics[width=0.95\textwidth]{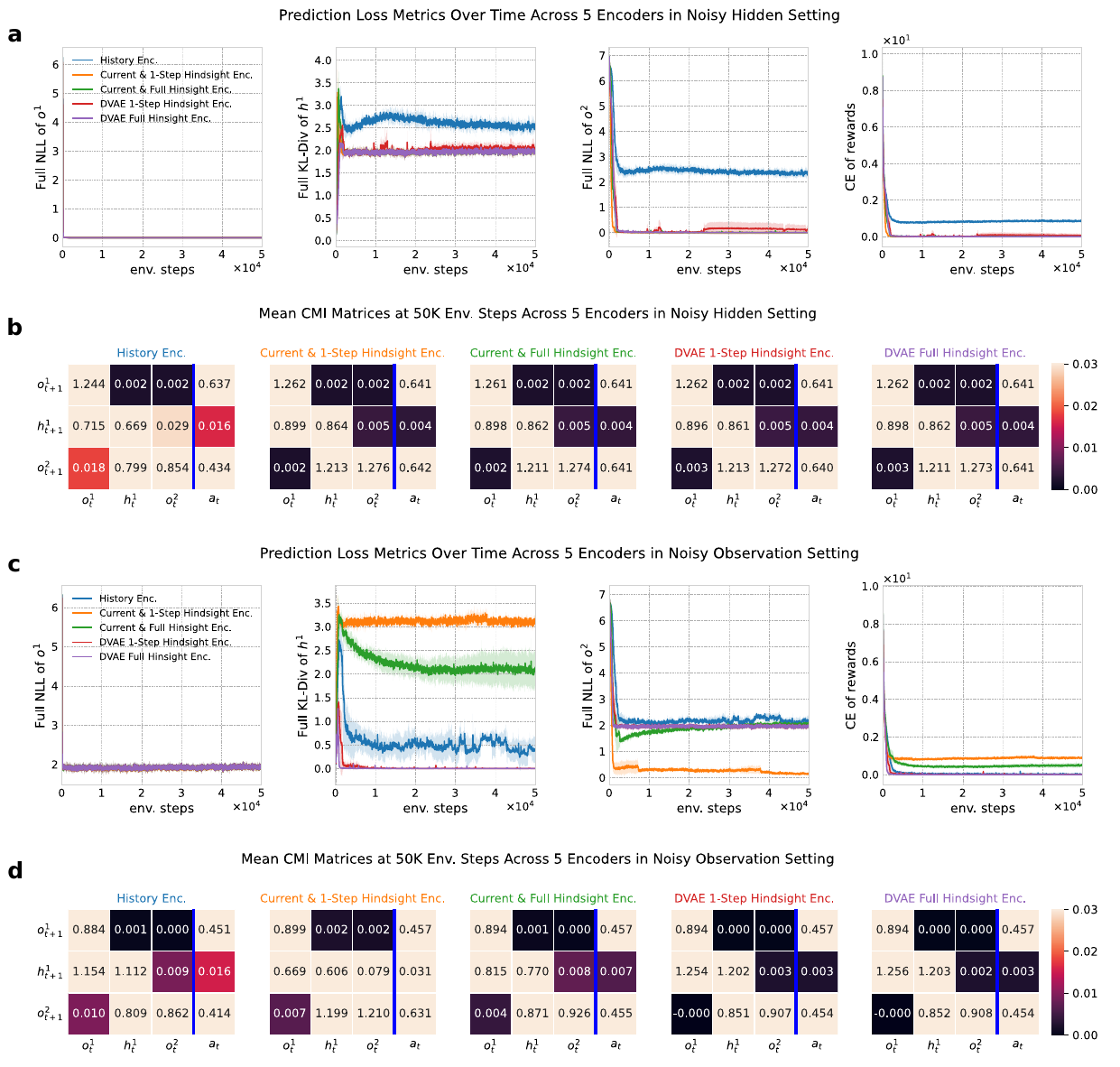}
\caption{\textbf{(a)} Comparison of 5 encoder types, showing training profiles of state and reward prediction losses with mean and standard deviation (std) when noise $\epsilon^h_t$ is applied to the hidden state transition. \textbf{(b)} Corresponding mean CMI matrices evaluated at the end of training across the 5 encoders, displayed as heatmaps under the same conditions. Similarly, \textbf{(c)} and \textbf{(d)} present training performance and evaluated CMI matrices, respectively, when noise $\epsilon^h_t$ is applied to the observed state transitions. In all experiment results, each loss metric and CMI calculation for each encoder is run over 3 seeds. The color bar range is capped at the CMI threshold $\delta$, so that light color denotes an edge, and dark color no edge. The DVAE encoders produce CMI matrices whose binarized values match ground-truth.}
\label{fig:cmi_evo}\vspace*{-1em}
\end{figure}

\paragraph{Results: DVAE Hindsight Encoders outperform History or Current \& Hindsight Encoders.} In Fig. \ref{fig:cmi_evo}, we empirically compare the training performances and evaluated CMI matrices across 5 types of encoders under 2 settings with exogenous noise $\epsilon_t$ applied to either the hidden state transition (noisy hidden setting) or the observed state transition (noisy observation setting). 

In the noisy hidden setting (Fig. \ref{fig:cmi_evo}a and b), encoders with hindsight information converge to zero loss for both observed state predictions (the full NLL term of the VLB in Eq. \eqref{eq:vlb_final}) and reward prediction (the cross-entropy loss in Eq. \eqref{eq:rew_loss}). 
These encoders also infer correct transition graphs after binarizing their evaluated CMI matrices using the threshold $\delta$. 
In contrast, the history-based encoder struggles to train effectively, resulting in a CMI matrix with values close to $\delta$, which reflects less statistical confidence in the existence of corresponding causal edges.
Without access to the next observation $o_{t+1}$, the history-based encoder cannot deterministically infer the current hidden state $h_t$, given the unknown noise $\epsilon_{t-1}$ affecting the transition to $h_t$.
However, hindsight-based approaches can learn $h_t$ by utilizing information from observed states, which serve as children of the hidden state in the transition graph, thereby enabling accurate learning of the transition graph. 
Due to the unobserved exogenous noise injected into the hidden state transition, the transition model can only predict the next hidden state in distribution. 
As a result, prediction losses for the hidden state (measured by the full KL divergence between the encoded and predicted next hidden states in Eq. \eqref{eq:vlb_final}) do not decrease to zero for all encoders.

In the noisy observation setting (Fig. \ref{fig:cmi_evo}c and d), the DVAE-based encoder successfully learns hidden representations, allowing it to accurately predict the next hidden states and rewards, while the history-based encoder and current hindsight-based encoder fail to achieve similar performance (as seen in the second and fourth panels of Fig. \ref{fig:cmi_evo}c). In the third panel of Fig. \ref{fig:cmi_evo}c), it appears that Current and Hindsight Encoders achieve lower loss, but this is due to learning to copy $o^2_{t+1}$ to the hidden, as described for Table \ref{tab:1hidden}. Encoders other than DVAE-based encoders produce CMI matrices with values closer to threshold, or even infer spurious edges.
We hypothesize that the DVAE-based model's ability to identify the current hidden state $h_t$ stems from its recursive structure, which combines sample-based past (Markovian) information with future information.
In contrast, other encoders, which rely on a single directional view of observables along the trajectory, lack sufficient information to identify the current hidden state in the noisy observation setting.
Similar to the prediction of next state hidden in the noisy hidden setting, the transition model can only predict noisy observations in distribution.

\begin{table}[t!]
\centering
\setlength{\tabcolsep}{6pt} % Adjust column spacing
\renewcommand{\arraystretch}{1} % Adjust row spacing

\resizebox{\textwidth}{!}{%
\begin{tabular}{cc|ccccc}
    \toprule
    \multicolumn{2}{c|}{} & \multicolumn{5}{c}{Evaluation Accuracy in Noisy Hidden / Observation Setting} \\
    \midrule
    \texttt{\#} Past & \texttt{\#} Future & Graph & $h^1$ Decoding & $o^1$, $o^2$ Prediction & $h^1$ Prediction & Reward Prediction \\
    \midrule
   \multicolumn{7}{c}{\textbf{History-Based Encoder}} \\
    all & 0 
    & \colorbox{lavender}{$0.944_{(0.039)}$} / $1.000_{(0.000)}$
    & \colorbox{lavender}{$0.865_{(0.019)}$} / \colorbox{beige}{$0.971_{(0.021)}$} 
    & $1.000_{(0.000)}$, \colorbox{lavender}{$0.872_{(0.023)}$} / $0.915_{(0.017)}$, \colorbox{beige}{$0.876_{(0.012)}$}
    & \colorbox{lavender}{$0.850_{(0.029)}$} / \colorbox{beige}{$0.964_{(0.026)}$}
    & \colorbox{lavender}{$0.927_{(0.020)}$} / \colorbox{beige}{$0.997_{(0.002)}$} \\
    \midrule
    \multicolumn{7}{c}{\textbf{Current and Hindsight-Based Encoder}} \\
    0 & 1 
    & $1.000_{(0.000)}$ / \colorbox{beige}{$0.944_{(0.079)}$}
    & $1.000_{(0.000)}$ / \colorbox{beige}{$0.866_{(0.044)}$}
    & $1.000_{(0.000)}$, $1.000_{(0.000)}$ / $0.915_{(0.017)}$, \colorbox{beige}{$0.996_{(0.003)}$}
    & $0.899_{(0.009)}$ / \colorbox{beige}{$0.814_{(0.009)}$}
    & $1.000_{(0.000)}$ / \colorbox{beige}{$0.949_{(0.005)}$} \\
    0 & all 
    & $1.000_{(0.000)}$ / $1.000_{(0.000)}$
    & $1.000_{(0.000)}$ / \colorbox{beige}{$0.914_{(0.017)}$}
    & $1.000_{(0.000)}$, $1.000_{(0.000)}$ / $0.915_{(0.017)}$, \colorbox{beige}{$0.897_{(0.007)}$}
    & $0.899_{(0.009)}$ / \colorbox{beige}{$0.870_{(0.031)}$}
    & $1.000_{(0.000)}$ / \colorbox{beige}{$0.959_{(0.009)}$} \\
    \midrule
    \multicolumn{7}{c}{\textbf{DVAE-Based Hindsight Encoder}} \\
    all & 1 
    & $1.000_{(0.000)}$ / $1.000_{(0.000)}$
    & $1.000_{(0.000)}$ / $1.000_{(0.000)}$
    & $1.000_{(0.000)}$, $1.000_{(0.000)}$ / $0.915_{(0.017)}$, $0.905_{(0.009)}$
    & $0.895_{(0.009)}$ / $1.000_{(0.000)}$
    & $1.000_{(0.000)}$ / $1.000_{(0.000)}$ \\
    all & all 
    & $1.000_{(0.000)}$ / $1.000_{(0.000)}$
    & $1.000_{(0.000)}$ / $1.000_{(0.000)}$
    & $1.000_{(0.000)}$, $1.000_{(0.000)}$ / $0.915_{(0.017)}$, $0.905_{(0.009)}$
    & $0.899_{(0.009)}$ / $1.000_{(0.000)}$
    & $1.000_{(0.000)}$ / $1.000_{(0.000)}$ \\
    \bottomrule
\end{tabular}%
}
\caption{Evaluation accuracies across various metrics, including transition graph accuracy (measured by the match between inferred and ground truth edges), hidden state decoding accuracy (linear decoding accuracy of encoded hidden states to ground truth hidden states), observation prediction accuracy, hidden state prediction accuracy (measured by the match between predicted and encoded next hidden states), and reward prediction accuracy. These metrics are reported for 5 types of encoders utilizing different steps of past and future observables in both noisy hidden and noisy observation settings. Each accuracy value is presented as $\text{mean}_{\text{std}}$ over 3 runs. Lavender and beige highlights indicate suboptimal accuracy values for certain encoders in the noisy hidden and observation settings, respectively. Note that the DVAE-based encoder is labeled as using all past observables, as it estimates the 1-step past hidden state based on recursive hidden samples from the beginning of an episode, which requires all past observables.}
\label{tab:1hidden}\vspace*{-1em}
\end{table}

We also tabulate the accuracy of graph edges, decoding of encoded hidden, and state transitions, after convergence, of the five encoder architectures across both noise settings, in Table \ref{tab:1hidden}. In the noisy hidden setting, the lower accuracies of the history-based encoder, highlighted in lavender, indicate its inability to learn the hidden state and accurately perform the corresponding transition and reward predictions.
Ideally, the encoded hidden state should be linearly decodable to its ground truth value and deterministically predictive of the reward, as reflected by perfect $h^1$ decoding and reward prediction accuracy in all other encoders. Additionally, the expected $h^1$ prediction accuracy should be approximately $0.9$, accounting for the 10\% noise in the hidden transition, assuming both the encoded and predicted hidden states are optimally learned. Indeed, the mean $h^1$ prediction accuracy for all encoders, except the history-based one, is very close to $0.9$.
Similarly, in the noisy observation setting, the accuracies highlighted in beige indicate suboptimal encoding and prediction of the hidden states for the history-based and current hindsight-based encoders. Interestingly, for the current and hindsight-based encoder, the mean $o^2$ prediction accuracy exceeds the expected value of $0.9$ and approaches $1.0$ (see also third panel of Fig. \ref{fig:cmi_evo}c), suggesting that this encoder copies its input of the next noisy $o^2$ as the hidden state. This copying approach, however, trades off accuracy in $h^1$ and reward prediction compared to encoders that do not learn this inconsequential solution for the hidden state.
The DVAE-based encoders perform optimally in both noise settings. Notably, the DVAE 1-step Hindsight Encoder achieves the same optimal performance as the theoretically-derived DVAE Full Hindsight Encoder due to absence of cascaded hidden factors in our environment.
% We hypothesize that since only one state is hidden in our example, hindsight along with present and  Markovian past information is sufficient to uniquely encode the current hidden. However, for a transition graph with multiple connected hidden states, multi-step future information might be required.

% We note that even in the noisy $h_1$ case, the output of the encoder $h_t$ becomes determistically determined by the past, current and 1-step future, whereas the transition model outputs a probabilistic $h_{t+1}$. However, if we were to also encode the noise, we would get a deterministic transition knowing the noise via the reparameterization lemma. We test this by inserting a noise encoder as below....

% \subsection{Scalability}
% \label{subsec:scal}

\begin{figure}[t!]
\centering
\includegraphics[width=0.95\textwidth]{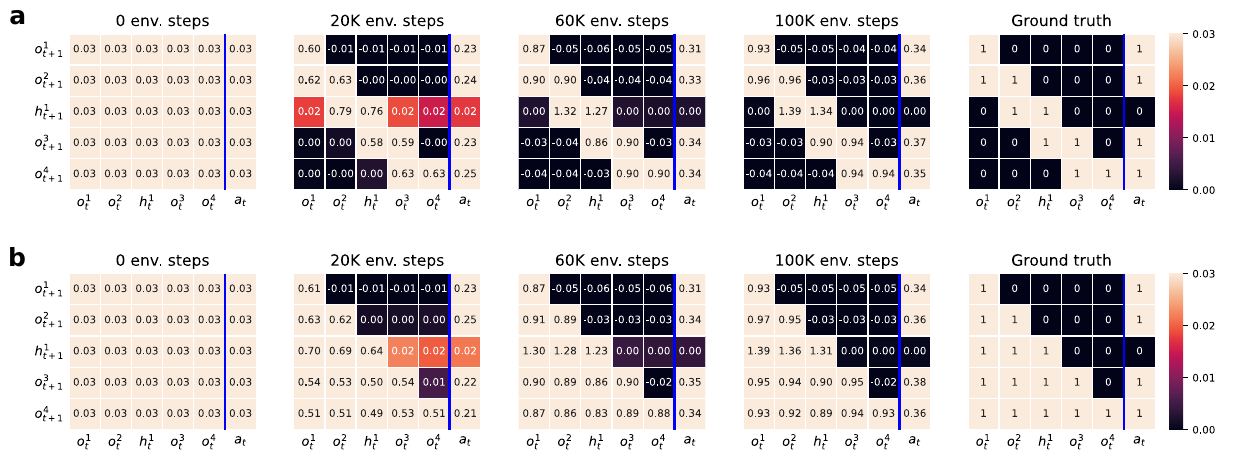}
\caption{Evolution of CMI matrices for the \textbf{(a)} chain and \textbf{(b)} full graph structures. The ground truth graphs are shown on the far right.}
\label{fig:general_g}\vspace*{-1em}
\end{figure}

Finally, we evaluate our DVAE Full Hindsight Encoder on chain and full structured (lower-triangular adjacency matrix) transition graphs with $d_S=5$, while keeping the rest of the environment setup unchanged.
Fig. \ref{fig:general_g} shows the evolution of the CMI matrix during training in the setting of noisy observations for both transition graph structures. The CMI matrix initially has all elements set to the predefined threshold $\delta$ and gradually decreases for unconnected factor pairs in the transition while increasing for connected factor pairs.
The final binary matrix, obtained by applying the threshold to binarize the CMI matrix, converges to the ground truth adjacency matrix.

\vspace*{-1em}\section{Discussion}
We have demonstrated that the proposed DVAE-based hindsight encoder effectively identifies hidden state factors and learns the causal transition graph in a factored-POMDP, outperforming both history-based and typical hindsight-based encoders. This approach shows particular promise in settings with access to full offline trajectories. In biological scenarios, our technique is reminiscent of ``trajectory replay" in rodent planning, where neural patterns associated with past experiences are replayed in both forward and reverse directions \citep{olafsdottir2018role}. Thus, our method holds value for applications where offline trajectories can be leveraged. In online settings, a causal model initially trained on offline trajectories could support more accurate model rollouts within frameworks like Dyna \citep{sutton1991dyna, sutton2012dyna, peng2018deep} or Model Predictive Control (MPC) \citep{chua2018deep, wang2019benchmarking, moerland2023model}, offering an advantage over models trained solely with history-based or other hindsight-based approaches.

In our formulation, we identified deterministic hidden components of factored state transitions, and, using the Reparametrization Lemma, isolated stochastic effects as unobserved exogenous noise per factor. Future work could refine our framework by also inferring the exogenous noise at each time step through dedicated noise encoders, following the identification of deterministic hidden factors. While our DVAE 1-step Hindsight Encoder was sufficient for a single hidden factor, extending it to scenarios with multiple cascaded hidden factors, with only the last hidden factor influencing an observable factor, may require additional future information for effective latent identification. Moreover, expanding this approach to continuous state-action spaces would link our work to DVAE research on latent dynamics in stochastic driven dynamical systems \citep{girin2020dynamical}. Addressing these areas would support further scaling and generalization of the framework.

\acks{C. Han, A. Gilra and E. Vasilaki acknowledge the CHIST-ERA grant for the ``Causal Explanations in Reinforcement Learning (CausalXRL)" project (CHIST-ERA-19-XAI-002), by the Engineering and Physical Sciences Research Council (EPSRC), United Kingdom (grant reference EP/V055720/1) for supporting the work. D. Basu acknowledges the CHIST-ERA grant for the CausalXRL project (CHIST-ERA-19-XAI-002) by L'Agence Nationale de la Recherche, France (grant reference ANR21-CHR4-0007), as well as the ANR JCJC for the REPUBLIC project (ANR-22-CE23-0003-01), and the PEPR project FOUNDRY (ANR23-PEIA-0003) for supporting the work. C. Han and E. Vasilaki acknowledge the grant for the ``Magnetic Architectures for Reservoir Computing Hardware (MARCH)" project, by the EPSRC, United Kingdom (grant reference EP/V006339/1) for supporting the work. C. Han, M. Mangan and E. Vasilaki acknowledge the grant for the ``Active learning and selective attention for robust, transparent and efficient AI (ActiveAI)" project, by the EPSRC, United Kingdom (grant reference EP/S030964/1) for supporting the work.}

\bibliography{clear_2025}
\clearpage
\appendix

\section{DVAE for Factored-POMDP}
\subsection{Log-likelihood decomposition}
\label{app:subsec:llh_decom}
The detailed derivation from Eq. \eqref{eq:mllh} to Eq. \eqref{eq:mllh_decom} is provided as follows:
\begin{align}
    \label{app:eq:mllh_decom_deri}
    & \mathbb{E}_{p(o_{1:T} | a_{1:T})} \left[\log p_\theta(o_{1:T} | a_{1:T})\right] \notag \\
    =& \mathbb{E}_{p(o_{1:T} | a_{1:T})} \left[\mathbb{E}_{q_\phi(h_{1:T} | o_{1:T}, a_{1:T})} \left[\log p_\theta(o_{1:T} | a_{1:T})\right] \right] \\
    =& \mathbb{E}_{p(o_{1:T} | a_{1:T})} \left[\mathbb{E}_{q_\phi(h_{1:T} | o_{1:T}, a_{1:T})} \left[\log \frac{p_\theta(o_{1:T}, h_{1:T} | a_{1:T})}{p_\theta(h_{1:T} | o_{1:T}, a_{1:T})} \right] \right] \\
    =& \mathbb{E}_{p(o_{1:T} | a_{1:T})} \left[\mathbb{E}_{q_\phi(h_{1:T} | o_{1:T}, a_{1:T})} \left[\log \frac{p_\theta(o_{1:T}, h_{1:T} | a_{1:T})}{q_\phi(h_{1:T} | o_{1:T}, a_{1:T})} \frac{q_\phi(h_{1:T} | o_{1:T}, a_{1:T})}{p_\theta(h_{1:T} | o_{1:T}, a_{1:T})}
    \right] \right] \\
    =& \mathbb{E}_{p(o_{1:T} | a_{1:T})} \left[\mathbb{E}_{q_\phi(h_{1:T} | o_{1:T}, a_{1:T})} \left[\log \frac{p_\theta(o_{1:T}, h_{1:T} | a_{1:T})}{q_\phi(h_{1:T} | o_{1:T}, a_{1:T})} \right] \right. \notag \\
    & \phantom{\mathbb{E}_{p(o_{1:T} | a_{1:T})} } + \left. \mathbb{E}_{q_\phi(h_{1:T} | o_{1:T}, a_{1:T})} \left[\log \frac{q_\phi(h_{1:T} | o_{1:T}, a_{1:T})}{p_\theta(h_{1:T} | o_{1:T}, a_{1:T})} \right] \right] \\
    =& \mathbb{E}_{p(o_{1:T} | a_{1:T})} \big[ \underbrace{\mathbb{E}_{q_\phi\left(h_{1: T} | o_{1: T}, a_{1: T}\right)} \left[ \log p_\theta\left(o_{1: T}, h_{1: T} | a_{1: T}\right) - \log q_\phi\left(h_{1: T} | o_{1: T}, a_{1: T}\right) \right]}_{\ell_{\mathrm{VLB}}(\theta, \phi ; o_{1: T}, a_{1: T}) } \notag \\
    & \phantom{\mathbb{E}_{p(o_{1:T} | a_{1:T})} } + D_{\mathrm {KL}}(q_\phi(h_{1:T} | o_{1:T}, a_{1:T}) \parallel p_{\theta}(h_{1:T} | o_{1:T}, a_{1:T})) \big]
\end{align}

\subsection{Variational Lower Bound (VLB) of DVAE-based framework for learning latent hiddens and transition dynamics of factored-POMDP}
\label{app:subsec:vlb_dvae}
\textbf{Generative model (transition model).} The generative model for the entire state sequence in the Eq. \eqref{eq:vlb_general} can be factorized as:
\begin{align}
    p_{\theta}(o_{1:T}, h_{1:T}|a_{1:T}) =& \prod_{t=0}^{T-1} p_{\theta}(o_{t+1}, h_{t+1}|o_{1:t}, h_{1:t}, a_{1:T}) \notag \\
    =& \prod_{t=0}^{T-1} p_{\theta_o}\left(o_{t+1}|o_{1:t}, h_{1:t+1}, a_{1:T}\right) p_{\theta_h}\left(h_{t+1}|o_{1:t}, h_{1:t}, a_{1:T}\right) \notag \\
    =& \prod_{t=0}^{T-1} p_{\theta_o}\left(o_{t+1}|o_{t}, h_{t}, a_{t}\right) p_{\theta_h}\left(h_{t+1}|o_{t}, h_{t}, a_{t}\right) 
    \label{eq:gen_causal}
\end{align}
where each term in the product is simplified using d-separation in the unrolled transition graph from $t=1$ to $T$ (see Fig. \ref{fig:graph_model}c). Here, $\theta=\theta_{o} \cup \theta_{h}$ represents the parameters of the generative model. Note that the observation likelihood $p_{\theta_o}\left(o_{t+1}|o_{t}, h_{t}, a_{t}\right)$ and the hidden prior $p_{\theta_h}\left(h_{t+1}|o_{t}, h_{t}, a_{t}\right)$ in the generative model corresponds to the transition models of the observed and hidden states, respectively.

\noindent\textbf{Inference model (hidden encoder).} Similarly, we factorize the posterior distribution of the generative model as follows:
\begin{align}
    p_{\theta}(h_{1:T}|o_{1:T}, a_{1:T}) 
    =& \prod_{t=0}^{T-1} p_{\theta}(h_{t+1}|h_{1:t}, o_{1:T}, a_{1:T}) \notag \\
    =& \prod_{t=0}^{T-1} p_{\theta}(h_{t+1}|h_{t}, o_{t:T}, a_{t:T}) 
    \label{eq:post_gen_dsep}
\end{align}
and consider that the inference model, parameterized by $\phi$, captures the exact factorized structure of the posterior distribution in Eq. \eqref{eq:post_gen_dsep}.
\begin{align}
    \label{eq:inf}
    q_{\phi}(h_{1:T}|o_{1:T}, a_{1:T}) = \prod_{t=0}^{T-1} q_{\phi}(h_{t+1}|h_{t}, o_{t:T}, a_{t:T})
\end{align}
Specifically, the hidden encoder $q_{\phi}(h_{t}|h_{t-1}, o_{t-1:T}, a_{t-1:T})$ combines information from the Markovian past, through $h_{t-1}$, $o_{t-1}$ and $a_{t-1}$, with information from the present and future observations $o_{t:T}$ and actions $a_{t:T}$ to encode the current hidden state $h_{t}$. It is important to note that we assume Markovianity for the forward transitions but not for the backward transitions. Consequently, the hidden encoder depends on all future information, rather than just the immediate next-step information used in the hindsight-based encoder by \citet{jarrett_curiosity_2023}.

\noindent\textbf{Variational Lower Bound.} By substituting the decomposed forms of both the generative model from Eq. \eqref{eq:gen_causal} and the inference model from Eq. \eqref{eq:inf} into the general form of VLB defined in Eq. \eqref{eq:vlb_general}, we obtain:
\begin{align}
    \label{eq:vlb_simplified}
    \ell_{\mathrm{VLB}}\left(\theta, \phi ; o_{1: T}, a_{1: T}\right) =& \sum_{t=0}^{T-1} \mathbb{E}_{q_\phi\left(h_{1: t} | o_{1: T}, a_{1: T}\right)}\left[\log p_{\theta_{o}}\left(o_{t+1} | o_{t}, h_{t}, a_{t}\right)\right. \notag \\
    &\left.- D_{\mathrm {KL}} \left(q_\phi\left(h_{t+1} | h_{t}, o_{t:T}, a_{t:T}\right)\right. \parallel \left.p_{\theta_{h}}\left(h_{t+1} | o_{t}, h_{t}, a_{t}\right)\right) \right]
\end{align}
By using the factorization in Eq. \eqref{eq:inf}, the expectation in the above VLB can be expressed as a cascade of expectations over conditional distributions of individual hidden states at different time steps:
\begin{align}
    \mathbb{E}_{q_\phi\left(h_{1:t} | o_{1:T}, a_{2:T}\right)}\left[f\left(h_t\right)\right] 
    =& \mathbb{E}_{q_\phi\left(h_1 | o_{1: T}, a_{1: T}\right)}[ \notag \\
    &\quad \mathbb{E}_{q_\phi\left(h_2 | h_1, o_{1: T}, a_{1: T}\right)}[ \notag \\
    &\quad \quad \mathbb{E}_{q_\phi\left(h_3 | h_2, o_{2: T}, a_{2: T}\right)}[ \ldots \notag \\
    &\quad \quad \quad \mathbb{E}_{q_\phi\left(h_t | h_{t-1}, o_{t-1: T}, a_{t-1: T}\right)} \left[f\left(h_t\right)\right] \ldots ]]]
\end{align}
Here, $f\left(h_t\right)$ represents an arbitrary function of $h_t$. Each intractable expectation in this sequence can be approximated using a Monte Carlo estimate. This involves iteratively sampling from \\ $q_\phi(h_\tau | h_{\tau-1}, o_{\tau-1: T}, a_{\tau-1: T})$ for $\tau=1$ to $t$, employing the same reparameterization trick used in standard VAEs \citep{maddison2016concrete, jang2016categorical, Kingma_2019}. Additionally, the VLB in Eq. \eqref{eq:vlb_simplified}, which is defined for a single data sequence, can be extended by averaging over a mini-batch of training data sequences, thereby approximating the expected VLB with respect to the true data distribution.

Furthermore, by expressing $o_{t} = (o_{t}^1, \ldots, o_{t}^{d_O})$, $h_{t} = (h_{t}^1, \ldots, o_{t}^{d_H})$ and $s_{t} = (o_{t}, h_{t})$ and using the factorized forms of both the transition models and the hidden encoder, we have:
\begin{align}
    \label{eq:model_factorized}
    p_{\theta_{o}}(o_{t+1} | o_{t}, h_{t}, a_{t}) =& \prod_{j=1}^{d_O} p_{\theta_{h}}(o_{t+1}^j | s_t, a_t), \\
    p_{\theta_{h}}(h_{t+1} | o_{t}, h_{t}, a_{t}) =& \prod_{j=1}^{d_H} p_{\theta_{h}}(h_{t+1}^j | s_t, a_t), \\
    q_{\phi}(h_{t+1} | h_{t}, o_{t:T}, a_{t:T}) =& \prod_{j=1}^{d_H} q_{\phi}(h_{t+1}^j | h_{t}, o_{t:T}, a_{t:T})
\end{align}

Eq. \eqref{eq:vlb_factorized} is obtained by substituting the above expressions into Eq. \eqref{eq:vlb_simplified}.

\subsection{Conditional mutual information}
\label{app:subsec:cmi}
Starting from the definition of conditional mutual information, we have:
\begin{align}
    I(s_t^i; s_{t+1}^j | s_{t} \backslash s_t^i, a_{t}) 
    =& \mathbb{E}_{p(s_{t}, a_{t}, s_{t+1}^j)} \left[ \log \frac{p_(s_{t}^i, s_{t+1}^j | s_{t} \backslash s_{t}^i, a_{t})}{p(s_{t}^i | s_{t} \backslash s_{t}^i, a_{t}) p(s_{t+1}^j | s_{t} \backslash s_{t}^i, a_{t})} \right] \label{eq:cmi_def} \\ 
    =& \mathbb{E}_{p(s_{t}, a_{t}, s_{t+1}^j)} \left[ \log \frac{p_(s_{t+1}^j | s_{t}, a_{t}) p_(s_{t}^i | s_{t} \backslash s_{t}^i, a_{t})}{p_(s_{t}^i | s_{t} \backslash s_{t}^i, a_{t}) p(s_{t+1}^j | s_{t} \backslash s_{t}^i, a_{t})} \right] \\ 
    =& \mathbb{E}_{p(s_{t}, a_{t}, s_{t+1}^j)} \left[ \log \frac{p_(s_{t+1}^j | s_{t}, a_{t})}{p(s_{t+1}^j | s_{t} \backslash s_{t}^i, a_{t})} \right] \label{eq:cmi_o} \\ 
    =& \mathbb{E}_{p(s_{t}, a_{t})} \left[ \mathbb{E}_{p(s_{t+1}^j | s_{t}, a_{t})} \left[ \log \frac{p_(s_{t+1}^j | s_{t}, a_{t})}{p(s_{t+1}^j | s_{t} \backslash s_{t}^i, a_{t})} \right] \right] \\
    =& \mathbb{E}_{p(s_{t}, a_{t})}\left[ D_{\mathrm {KL}} ( p(h_{t+1}^j | s_{t}, a_{t}) \parallel p(h_{t+1}^j | s_{t} \backslash s_{t}^i, a_{t}) ) \label{eq:cmi_h} \right]
\end{align}
where Eqs. \eqref{eq:cmi_o} and \eqref{eq:cmi_h} correspond to Eqs. \eqref{eqn:CMI_o} and \eqref{eqn:CMI_h}, respectively.

\subsection{Neural Network-Based Parameterization}
\label{app:subsec:nn_paramz}
The hidden encoder $q_\phi\left( h_{t}|h_{t-1}, o_{t-1:T}, a_{t-1:T} \right)$ is implemented using a backward RNN to capture current and future dependencies, and an MLP to model Markovian past dependencies. A combiner function (CF) is then employed to merge the outputs of the MLP and the RNN (its internal state) to produce parameters  (e.g., logits) of the distribution of the current hidden state:
\begin{align}
    \overleftarrow{g}_t & =\mathrm{RNN}_{\phi_{\overleftarrow{g}}}\left(\overleftarrow{g}_{t+1}, \left[o_t, a_{t}\right]\right), \\
    e_t & =\mathrm{MLP}_{\phi_{e}}\left(h_{t-1}, o_{t-1}, a_{t-1}\right), \\
    f_t & =\mathrm{CF}_{\phi_{f}}\left(e_{t}, \overleftarrow{g}_t\right), \\
    q_\phi\left( h_{t}|h_{t-1}, o_{t-1:T}, a_{t-1:T} \right) & =\mathrm{Dist}\left(h_t ; f_t\right)
\end{align}
where $\mathrm{CF}_{\phi_{f}}$ is a feedforward combining network parameterized by ${\phi_{f}}$. Thus, the parameters of the hidden encoder are $\phi=\phi_{\overleftarrow{g}} \cup \phi_{e} \cup \phi_{f}$.

The transition model for the observed states $p_{\theta_o}\left(o_{t+1}|o_{t}, h_{t}, a_{t}\right)$ and the hidden states $p_{\theta_h}\left(h_{t+1}|o_{t}, h_{t}, a_{t}\right)$ are implemented using factor-wise masked MLPs (MMLPs) following \cite{wang2022causal}:
\begin{align}
    m_t & =\mathrm{MMLP}_{\theta_{o}}\left(o_{t}, h_{t}, a_{t}\right), \\ p_{\theta_o}\left(o_{t+1}|o_{t}, h_{t}, a_{t}\right) & =\mathrm{Dist}\left(o_t ; m_t\right), \\
    n_t & =\mathrm{MMLP}_{\theta_{h}}\left(o_{t}, h_{t}, a_{t}\right), \\ p_{\theta_h}\left(h_{t+1}|o_{t}, h_{t}, a_{t}\right) & =\mathrm{Dist}\left(h_t ; n_t\right) 
\end{align}
where $m_t$ and $n_t$ are the outputs of the masked MLPs parameterized by $\theta_o$ and $\theta_h$, respectively. The distributions $\mathrm{Dist}\left(o_t ; m_t\right)$ and $\mathrm{Dist}\left(h_t ; n_t\right)$ represent the probability distributions of  $o_{t+1}$ and $h_{t+1}$ parameterized by $m_t$ and $n_t$.

\begin{figure}[ht]
\centering
\includegraphics[width=0.8\textwidth]{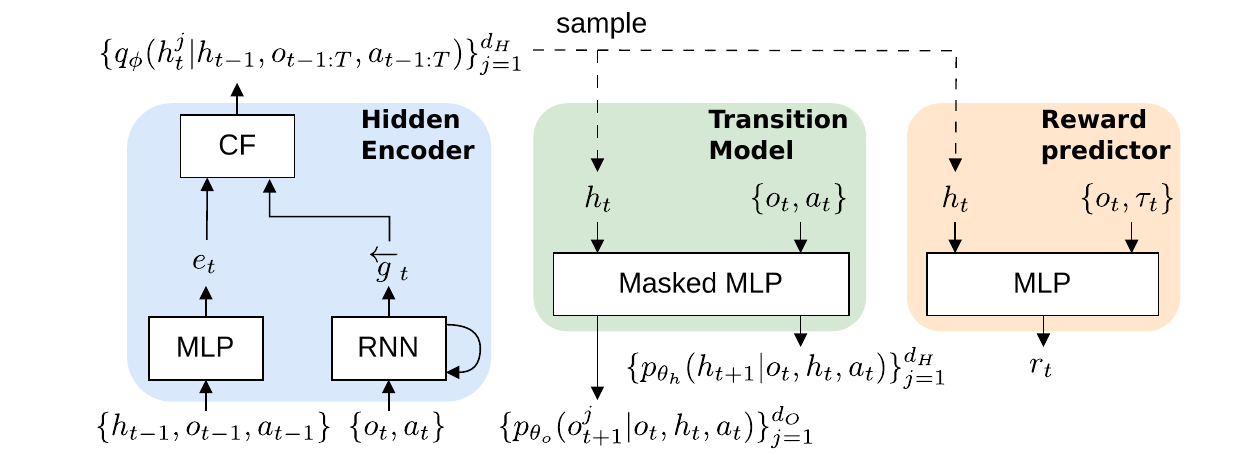}
\caption{Model architecture illustrating the computational graph for encoding, sampling and prediction processes.}
\label{fig:dvae_arch}
\end{figure}

The architecture of the DVAE model is illustrated in Fig. \ref{fig:dvae_arch}.

\iffalse
\begin{align*}
    P(h_{t}|o_{t:t+1},a_{t})
    P(h_{t}|h_{t-1},o_{t-1:T},a_{t-1:T})
\end{align*}

$Z = h_t$, $X=o_{t:t+1},a_{t}$, $Y=h_{t-1}, o_{t-1}, a_{t-1}, o_{t+2:T},a_{t+1:T} = pa \cup fu$

\begin{align*}
    &P(h_{t-1}, o_{t-1}, a_{t-1}, o_{t+2:T},a_{t+1:T}|o_{t:t+1},a_{t}) \\
    &= P(h_{t-1}, o_{t+2:T},a_{t+1:T}|o_{t:t+1},a_{t}) P(o_{t-1}, o_{t+2:T},a_{t+1:T}|o_{t:t+1},a_{t}) P(a_{t-1}, o_{t+2:T},a_{t+1:T}|o_{t:t+1},a_{t})\\
     &= P(h_{t-1},|o_{t:T},a_{t:T}) P(o_{t-1}|o_{t:T},a_{t:T}) P(a_{t-1}|o_{t:T},a_{t:T}) P^3(o_{t+2:T},a_{t+1:T}|o_{t:t+1},a_{t}) \\
     &=P(pa|o_{t:t+1},a_{t},fu) P(fu|o_{t:t+1},a_{t})
\end{align*}

\begin{align*}
    P(Z|X,Y) = P(Z,Y|X)/P(Y|X) 
\end{align*}
\begin{align*}
    P(Z|X) = \int P(Z|X,Y) P(Y|X) \leq \max_Y P(Z|X,Y)
\end{align*}

\begin{align*}
    \frac{P(Z|X,Y)}{\max_Y P(Z|X,Y)} \leq \frac{P(Z|X,Y)}{P(Z|X)} = \frac{P(Z|X,Y)}{\int P(Z|X,Y) P(Y|X)} = \frac{P(Z|X,Y)P(Y|X)}{\int P(Z|X,Y) P(Y|X)}\frac{1}{P(Y|X)} \leq 1/P(Y|X)
\end{align*}

\begin{align*}
    I(P(Z);P(X))\leq I(P(Z,Y);P(X,Y)) = \int_z \int_y 
\end{align*}
We have a markov chain $pa \rightarrow pre \rightarrow fu$
\begin{align*}
    I(pre; pa, fu) = I(pre; pa) + I(pre;fu|pa) = I(pre;fu) + \underset{=0}{I(pre;pa|fu)}
\end{align*}
\fi
\end{document}